# Deep Latent Factor Model for Collaborative Filtering

Aanchal Mongia[1], Neha Jhamb[1], Emilie Chouzenoux[2] and Angshul Majumdar[1]

[1]Indraprastha Institute of InformationTechnology, New Delhi, India 110020

[2]University of Paris-Est, LIGM, UMR CNRS 8019, France

aanchalm@iiitd.ac.in, neha16037@iiitd.ac.in, emilie.chouzenoux@univ-mlv.fr and angshul@iiitd.ac.in

**Abstract**— Latent factor models have been used widely in collaborative filtering based recommender systems. In recent years, deep learning has been successful in solving a wide variety of machine learning problems. Motivated by the success of deep learning, we propose a deeper version of latent factor model. Experiments on benchmark datasets shows that our proposed technique significantly outperforms all state-of-the-art collaborative filtering techniques.

## 1.  Introduction

Today much of retail business is online. Retail ecommerce is having such an impact on economy that economists are dubbing it as the 'Amazon effect'. Ecommerce is generating more jobs and checking inflation rates (at least in the USA). Most of its success is owing to recommender systems.

Unlike physical stores, ecommerce portals deal with literally millions of products. It is not possible for the customer / user to sift through all the products in a finite amount of time to find out what he / she needs. The users rely on the suggestions from the recommender system to zero-in on products they like.

The benefit of recommender system goes in both ways. If the recommendations are not good, the user may stop relying on its suggestions and stop using the ecommerce portal. This harms the user – as he / she does not get what is required; it is also detrimental to the portal – as it loses out on revenue. To improve customer satisfaction, goodwill and revenue, it is thus mandatory for the ecommerce portals to have a very accurate recommendation system.

In the initial days of recommender systems, classical information retrieval based approaches such as content based filtering were used. However, such techniques relied on expert opinions and were very sector specific (fashion and books would be treated completely differently) and hard to generalize as algorithms. This led to its early demise. For the last two decades collaborative filtering has been the de facto approach behind recommender systems.

Collaborative filtering can be categorized into several branches. The initial approach was based on neighbourhood based models. They were intuitive, easy to understand and implement but lacked in accuracy. The second approach was based on classification. This yields slightly better results but is very hard to explain. The third, and the most prominent approach today is based on latent factor models. This class of techniques is more abstract and requires good understanding of mathematics for interpretation, but are nevertheless more accurate.

In the last half a decade deep learning has made inroads into almost all aspects of applied computer science – speech processing, computer vision, NLP etc. It has also seen some applications in information retrieval.

Motivated by the success of deep learning, we propose a deep latent factor model. This would be a new deep learning tool, specifically tailored for collaborative filtering problems.

The shallow / standard latent factor model was based on the matrix factorization approach. The advent of deep learning, generalized matrix factorization to deeper versions. In this work, we follow the same idea and extend the latent factor model to a deeper version. However, it must be noted that the extension from deep matrix factorization techniques to our deep latent factor model is non-trivial. In the former, all the observations are known but in our case, only a partial set of observations are known – this makes the problem more challenging.

## 2. Literature Review
### 2.1 Neighborhood / Similarity Based Models

Neighborhood based models try mimicking the nature of human interaction. It finds out users having similar taste as that of the active user. Among these similar users, weight is given according to the similarity; i.e. more the similarity more the weight. The rating on a particular item from the similar users is multiplied by these weights to estimate the rating from the active user.

To find out the active user's rating on a particular item, the ratings of the similar users are interpolated. This is given by,

$$v_{a,j} = \sum_{i \in neighbours(a)} w_i v_{i,j} \tag{1}$$

Here $a$ is the active user; we want to predict the missing rating of user $a$ for the item $j$. For this, we look at all the neighbours of $a$, indexed by $i$ and interpolate their rating by multiplying it by the linear interpolation weights $w_i$. Such a technique is called user-user recommendation [1, 2].

The way we have looked at collaborative filtering from the user's perspective, we can also look at it from the items perspective. The two are exactly similar. This leads to the item-item collaborative filtering [3, 4]. Here instead of trying to find similar users one has to find similar items. The score of a user on the active item is found by interpolating the ratings given the same user on similar items. The interpolation weights, as previously are usually based on normalized similarity measure.

There are many variants to these basic approaches. Some studies combine the item and the user-based models [5]. Other studies have posed such similarity based collaborative filtering as a graph signal processing problem [6]; however the basic approach remains the same there in. Such techniques are simple to understand and analyze. They cannot compete with more recent abstract mathematical models, but may be preferred by practitioners for the ease of implementation and understanding. Since neighborhood based models is not the topic of our interest, we do not discuss it any further. The interested reader can peruse [7].

### 2.2 Latent Factor Models

In content based filtering, one had to exclusively specify the factors that were thought responsible for user's choice on a particular class of items. As a result, the algorithms were not generalizable; the factors responsible for footwear were different from the ones responsible for selecting for instance movies. That is the primary reason for its failure.

Latent factor model [8-10] is a generalization of content based filtering. It is based on the assumption that the factors responsible for the user's choice on an item need not to be explicitly known. A user $i$ can be defined by its affinity towards these latent factors, represented by $u_i$ and an item $j$ can be defined by its corresponding latent factor $v_j$. The rating is high when the two latent factors match (same as content based filtering). This is best modeled by the inner product between the user's and item's latent factors. The rating of the $i^{th}$ user on the $j^{th}$ item is modeled as:

$$x_{i,j} = u_i v_j, \forall\ i, j \tag{2}$$

Considering the entire ratings matrix for M users and N items (2) can be represented as:

$$X = UV \text{ where } U = [u_1 | ... | u_M]\ V^T = [v_1 | ... | v_N]$$

Had the full ratings matrix be known, the problem would be trivial. What makes it challenging is the fact that the matrix is only partially observed; the goal is to infer the missing ratings. Once this is done, one can suggest items to users with high predicted ratings. Mathematically we can express it as:

$$Y = R \cdot X = R \cdot (UV) \tag{3}$$

Here R is a binary sampling mask consisting of 0s where ratings are missing and 1s where they are present; symbol '·' indicates element-wise product.

One can estimate the latent factor matrices for the users and the items by solving the following problem.

$$\min_{U,V} \|Y - R \cdot (UV)\|_F^2 + \lambda \left( \|U\|_F^2 + \|V\|_F^2 \right) \tag{4}$$

The ridge regression type penalties are there to check over-fitting. There are many algorithms for solving (4) starting from simple alternative least squares to multiplicative updates to block conjugate gradients. For all practical purposes, the simple alternating least squares yields good results.

The factorization problem (4) is bi-linear in the variables. Therefore, there is no guarantee of convergence to a global minima; results exists only for convergence to local minima. This issue can be overcome by directly solving for the ratings themselves instead of the latent factors.

The assumption here is that the ratings matrix is of low-rank. This follows directly from the latent factor model; the rank of the matrix is the same as the number of factors. The most direct way to solve for the ratings would be to find an X that has the lowest rank. But the rank minimization is known to be NP hard and hence there is no tractable solution (every algorithm is as good as brute force search). To ameliorate this problem theoretical studies [11-13] have shown that one can guarantee a minimum rank solution (under certain circumstances) by relaxing the NP hard rank minimization problem to its closest convex surrogate – the nuclear norm. Mathematically this is expressed as,

$$\min_X \|Y - R \cdot X\|_F^2 + \lambda \|X\|_{NN} \tag{5}$$

Here $\|.\|_{NN}$ denotes the nuclear norm; defined as the sum of singular values. This (5) is a convex problem that can be solved by semi-definite programming. Today more efficient algorithms exist [14].

The number of entries in Y are far fewer than the number of entries in X; for academic problems only 5% of the data is available and in practice less than 1% of the data is available. This makes collaborative filtering a highly under-determined problem. In such a scenario, any secondary information is likely to improve the results. For example, studies like [15, 16] have shown that using the associated metadata (user's demographic and item's metadata) can indeed improve recommendation accuracy. Other studies [17, 18] showed that by harnessing the power of neighborhood based models into the latent factor based method can also improve the results. However in this work, we are not concerned about using associated information, and hence these techniques will not be discussed any further.

## 2.3 Representation Learning

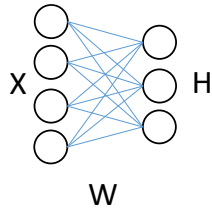

**Fig. 1.** Restricted Boltzmann Machine

Restricted Boltzmann machine (RBM) [19, 20] (Fig. 1) is popular today as a building block for deep belief network; but it was originally introduced for solving the collaborative filtering problem. However, owing to its inherent restrictions, foremost among them being the constraint on the input to be 1 or 0, RBMs never became popular in the context of collaborative filtering. Neither are the modified Gaussian Bernoulli RBMs, which expect continuous valued inputs in the range between 0 to 1, suitable for such inputs. Drawing similarity with the latent factor models, one can see that the network weights can be seen as users' latent factors and the representation as items' latent factors. The RBM is learnt by contrastive divergence. This is a cumbersome technique not amenable to mathematical manipulations.

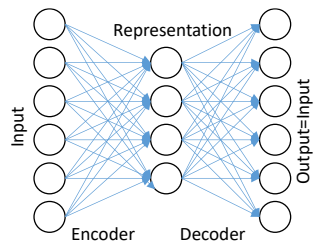

**Fig. 2.** Autoencoder

In recent times, autoencoder based collaborative filtering techniques have shown promising results [21-26]. They rely on a typical neural network where the output is the same as the input. The schematic diagram is shown in Fig. 2. When used for collaborative filtering, the input (and output) has missing values. When there are such missing entries in the data, the corresponding network weights are not updated. Once the training is finished, the representation from the encoder is multiplied by the decoder to get the full ratings matrix.

Deep / stacked autoencoders have also been used for collaborative filtering. Conceptually, the input and the outputs remain the same. The only difference between the shallow (Fig. 2) and deep autoencoeder is that the later has multiple layers of encoders and decoders. However, there is no significant gain in going deeper as was shown in [23, 24].

In recent years, with the success of deep learning in almost all areas of applied machine learning, such techniques have been leveraged for collaborative filtering as well; see for instance [23 – 26].

Most other studies in deep learning based collaborative filtering are somewhat heuristic. For example in [27], the inputs to the deep neural network are simply the IDs of the user and the item and the output is the corresponding rating. Such a model is likely to be arbitrary since the ID of the user and the item does not carry information about either. Therefore it does not make any logical sense to predict rating from such inputs.

The next work [28] is more sensible. Instead of using user's and item's IDs as inputs, it characterizes each user by his/her ratings on all items and characterizes each item by all its available ratings. This too uses a classification based framework where the output is the corresponding rating of the user on the item. Though a more logical approach than [26], using ratings for both inputs (user and item) and as the output (class) is difficult to justify. Unsurprisingly that neither [26] nor [27] provides any justification of their model.

The most thorough and logical approach to deep learning based collaborative filtering has been proposed in [28]. It is a deep neural network based regression framework. It uses user's past history (of item ratings) as input and the top recommended new item as output. Note that it does not perform collaborative filtering in the conventional sense since it does not predict ratings.

## 3. Proposed Deep Latent Factor Model
### 3.1 Latent Factor Model as Neural Network

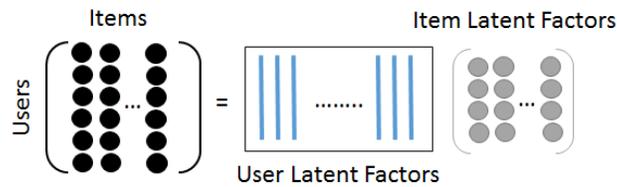

**Fig. 3.** Latent Factor Model

Fig. 3 depicts the standard latent factor model. We have assumed users to be along the rows and items along the columns. The entire ratings matrix is expressed as a product of user latent factor matrix and item latent factor matrix. Being a product of two matrices it is expressed mathematically in the form of matrix factorization.

In this work we look at the latent factor model as a neural network. Instead of looking at the user latent factors as vectors, we can think of them as connections from the item latent factors to the ratings. This is shown in Fig. 4.

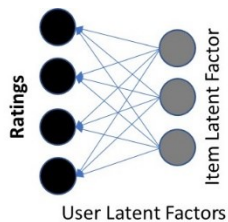

**Fig. 4.** Neural Network Interpretation

Once we have the neural network type interpretation of collaborative filtering, it is easy to conceptualize the deeper extensions. In any deep learning architecture, the latent representation from one layer acts as input to the subsequent layer. We follow the same principle in proposing the deep latent factor model.

## 3.2 Deep Latent Factor Model

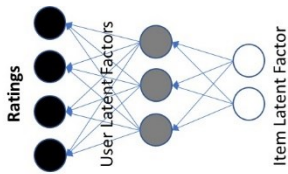

**Fig. 5.** Deep Latent Factor Model

A deeper (2-level) latent factor model is shown in Figure 5. This can be extended to deeper layers. Like all other deep learning tools, we suffer from the limitations of mathematical interpretability. Intuitively speaking, as one goes deeper, more abstract representations are observed. For example the latent factor model stems from the assumption that there are only a few factors that guide or decision in choosing a book or a movie. For a book, it may be just the name of the author, or the genre or the publisher. For movies the number of factors may be slightly more – it can be the star cast, the director, the main technical crew. It is not possible to capture all the factors objectively, which justifies the latent factor model. In a deeper latent factor model, instead of capturing the superficial factors (author, actor etc.) it is likely that deeper personality traits are captured. For example the deeper nodes may refer to five classic personality traits [26, 27] – *introversion, extroversion, neuroticism, openness* and *conscience*. Expressing the models based on such abstract fundamental traits is likely to make it more robust and accurate.

The neural network type interpretation is not mandatory, but it is easier to align with other deep learning methods. One can express the deep latent factor model in the more usual matrix factorization form (Fig. 6).

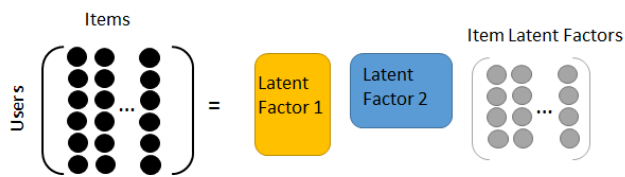

**Fig. 6.** Factorization Type Interpretation

*Mathematical Formulation*

In the standard latent factor model the ratings matrix is expressed as a product of user and item latent factors.

$$X = UV \qquad (6)$$

Usually non-negativity is enforced on the latent factors, in a similar manner than in the rectified linear unit (ReLU) type activation function in neural network.

In our deeper model, we will have multiple levels of user latent factors and one final level of item latent factors. This can be expressed as,

$$X = U_1 U_2 V \qquad (7)$$

Here $U_1$ and $U_2$ are the two levels of item latent factors and $V$ consists of the deep user latent factors. Such a deep factorization model has been proposed before [29, 30]. In each level, a ReLU type activation was imposed by non-negativity constraints. The formulation can be easily extended to more layers.

$$X = U_1 U_2 ... U_N V \qquad (8)$$

Techniques for solving deep matrix factorization have already been developed [29, 30] when the data is fully observed. Unfortunately in collaborative filtering, the ratings matrix is only partially observed.

$$Y = R \cdot X \qquad (9)$$

where $X$ is the full ratings matrix, $R$ the binary matrix of 1's and 0's and $Y$ the acquired ratings matrix.

We incorporate the deep matrix factorization framework (7) into the partially observed model (8) to arrive at our deep latent factor model.

$$Y = R \cdot X = R \cdot (U_1 U_2 ... U_N V) \qquad (10)$$

*Solution*

We will show the algorithm for three layers; it will be generic enough for more or fewer layers. The formulation for three layers is –

$$\min_{U_1, U_2, U_3, V} \tfrac{1}{2} \| Y - R \cdot (U_1 U_2 U_3 V) \|_F^2 \text{ such that}$$
$$U_1 U_2 U_3 V \geq 0, U_1 \geq 0, U_2 \geq 0, U_3 \geq 0, V \geq 0 \qquad (11)$$

This is equivalent to the following,

$$\min_{U_1, U_2, U_3, V, X} \tfrac{1}{2} \| Y - R \cdot (U_1 U_2 U_3 V) \|_F^2 \qquad (12)$$

such that $U_1 U_2 U_3 V = X$ and $X \geq 0, U_1 \geq 0, U_2 \geq 0, U_3 \geq 0, V \geq 0$. We propose a projected gradient method [31], with inner loop based on alternating projection [32] to solve (11). The general form of the algorithm is as follows:

Initialize: $X^0, U_1^0, U_2^0, U_3^0, V^0$

For $k = 1, 2, \ldots$

1. $\bar{X}^k = X^k - \gamma \left( R^T \cdot (R \cdot X^k - Y) \right)$

2. $\left( X^{k+1}, U_1^{k+1}, U_2^{k+1}, U_3^{k+1}, V^{k+1} \right)$ solution of

$$\min_{X, U_1, U_2, U_3, V} \left\| \bar{X}^k - X \right\|_F^2 \text{ s.t. } U_1 U_2 U_3 V = X$$

and $X \geq 0, U_1 \geq 0, U_2 \geq 0, U_3 \geq 0, V \geq 0$

The second sub-problem corresponds to projecting $X^k$
on the constrained domains $U_1 U_2 U_3 V = X$
and $X \geq 0, U_1 \geq 0, U_2 \geq 0, U_3 \geq 0, V \geq 0$.
This projection, can be solved by an inner loop
based on the following alternating projections

$X^{k+1} = P_+ \left( \bar{X}^k \right)$

$U_1^{k+1} = P_+ \left( U_1^k - \left( U_1^k U_2^k U_3^k V^k - X^{k+1} \right) \left( U_2^k U_3^k V^k \right)^\dagger \right)$

$U_2^{k+1} = P_+ \left( U_2^k - \left( U_1^{k+1} \right)^\dagger \left( U_1^{k+1} U_2^k U_3^k V^k - X^{k+1} \right) \left( U_3^k V^k \right)^\dagger \right)$

$U_3^{k+1} = P_+ \left( U_3^k - \left( U_1^{k+1} U_2^{k+1} \right)^\dagger \left( U_1^{k+1} U_2^{k+1} U_3^k V^k - X^{k+1} \right) \left( V^k \right)^\dagger \right)$

$V^{k+1} = P_+ \left( V^k - \left( U_1^{k+1} U_2^{k+1} U_3^{k+1} \right)^\dagger \left( U_1^{k+1} U_2^{k+1} U_3^{k+1} V^k - X^{k+1} \right) \right)$

Here we have used the general property that the projection of a matrix $U$ on a constraint $X = AUB$ is given by

$$\hat{U} = U - A^\dagger (AUB - X) B^\dagger \tag{13}$$

Steps 1 and 2 can be understood as a gradient projection method, with stepsize $\gamma$. The complicated form of the constraint requires the use of an inner loop for the projection step. Here, we have proposed to make use of alternating projections. The convergence of the whole scheme cannot be established easily, because of the presence of the coupling in the equality constraint $U_1 U_2 U_3 V = X$. However, in the experimental results, we will show that the algorithm converges empirically.

A recent work [33], proposes a hierarchical latent factor model. Unlike ours, which just goes deep, [33] bifurcates in each layer. The resulting model increases complexity since there are more variables to solve now and extra constraints to impose the tree structure. This is likely to make the model more vulnerable to over-fitting. The work claims convergence, however they provide no insights into this claim.

*Computational Complexity*

The computational cost of the proposed method is governed by the pseudo-inverse operations that must be performed in Step 2. The complexity of it is given by $O(n^w)$ where w<2.37 and is conjectured to be 2. Note however that one can play with the size of the latent factors matrices, in order to control the memory and computational burden of these inversions. An alternative for very large datasets is to use conjugate gradient solver, but in our practical experiments, it appears not necessary.

# 4. Experimental Evaluation
## 4.1 Datasets

We carry our evaluation on movie recommendations. Experiments are carried out on three standard datasets – Movielens 100K, Movielens 1M, Movielens and 10M. All of them are from https://grouplens.org/datasets/movielens/.

  (1) movie-100K: 100,000 ratings for 1682 movies by 943 users;

  (2) movie-1M: 1 million ratings for 3900 movies by 6040 users;

  (3) movie-10M: 10 million ratings for 10681 movies by 71567 users.

For these datasets the splits are pre-defined. The protocol is to carry out 5 fold cross-validation on these sets.

## 4.2 Convergence

We show the empirical convergence of our algorithm on the first two datasets. One can see that our algorithm decreases the cost function monotonically. The result on the third (10M) dataset is similar; hence we do not show it.

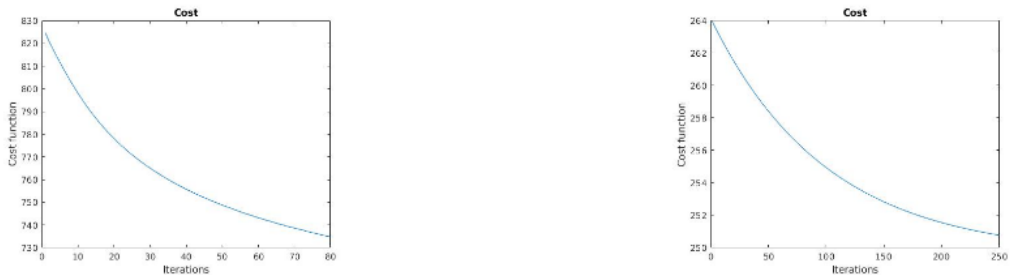

**Fig. 6.** Convergence plot. Left – 100K and Right – 1M.

## 4.3 Comparative Results

We have compared our technique with some state-of-the-art deep methods – collaborative deep learning (CDL) [25], marginalised deep autoencoder (MDA) [24] and deep matrix factorization (DMF) [28]. We also compare with a recent shallow yet robust approach called robust matrix factorization (RMF) [34]; although it has never been used for collaborative filtering, it surpasses standard matrix factorization for other tasks. We have compared with the hierarchical latent factor model (HLFM) [33].

For our proposed method we have shown the results for two and three levels. Since the number of users / items range in 1000's this is as deep as we can go. The two level architecture is 40-20 and the three level architecture is 40-20-10. The reason for choosing 40 latent factors in the first level is fairly well known. Usually for movie recommendations 40 is a good number of latent factors. For deeper layers we use the usual rule of thumb where the number of factors are reduced by half in every layer. We tried going deeper, but the results deteriorated; so we do not show those results here. Our method just requires one parameter to be specified; we have used γ=0.1 throughout. We have found that our algorithm is robust to the choice of parameter between 0.01 to 0.8. Outside the said range, the performance degrades gracefully.

For initialization we follow an deterministic SVD based decomposition. The missing entries are first filled by the row and column averages. SVD is performed on the thus filled matrix. The left singular vectors are used for initializing the $U_1$. Then SVD is performed on the remaining portion, i.e. the product of singular values and the right singular vectors. The left singular vectors of the second SVD is used to initialize $U_2$. This process is continued. For the final layer, the product of the singular values and the right singular vectors is used to initialize V.

One must note that the methods compared against (MDA, CDL, DMF, RMF and HLFM) have not been used on standard protocols defined on all these datasets. Therefore, we have followed the approach presented in the corresponding papers to tune the parameters for our datasets.

As the evaluation metric, we show results on all the standard ones – mean absolute error (MAE), root mean squared error (RMSE), precision and recall. The results are shown in the following tables (I to III). The results from our method are shown in bold.

One can see that we improve upon the rest in terms of every possible metric; the improvements are considerably large. To put the results in perspective, one must remember that the Netflix prize of 1 million was given to the winners who reduced the RMSE on the Netflix dataset from 0.95 to 0.85. We do not give results on the Netflix dataset owing to concerns over a pending lawsuit on the usage of the dataset.

TABLE 1. Results on 100K Movielens

| Method | MAE | RMSE | Precision | | Recall | |
|---|---|---|---|---|---|---|
| | | | @10 | @20 | @10 | @20 |
| RMF | .732 | .938 | .515 | .380 | .641 | .767 |
| DMF | .735 | .940 | .513 | .380 | .641 | .775 |
| CDL | .742 | .953 | .510 | .377 | .642 | .765 |
| MDA | .758 | .981 | .513 | .372 | .641 | .767 |
| HLFM | .750 | .962 | .512 | .376 | .640 | .769 |
| **40-20** | **.726** | **.939** | **.522** | **.380** | **.649** | **.782** |
| **40-20-10** | **.717** | **.901** | **.536** | **.399** | **.658** | **.792** |

TABLE 2. Results on 1M Movielens

| Method | MAE | RMSE | Precision | | Recall | |
|---|---|---|---|---|---|---|
| | | | @10 | @20 | @10 | @20 |
| RMF | .689 | .876 | .669 | .526 | .625 | .792 |
| DMF | .691 | .878 | .671 | .523 | .625 | .799 |
| CDL | .689 | .871 | .671 | .531 | .630 | .802 |
| MDA | .686 | .879 | .669 | .526 | .625 | .791 |
| HLFM | .698 | .880 | .661 | .517 | .619 | .785 |
| **40-20** | **.681** | **.864** | **.673** | **.531** | **.636** | **.799** |
| **40-20-10** | **.678** | **.854** | **.691** | **.543** | **.641** | **.809** |

TABLE 3. Results on 10M Movielens

| Method | MAE | RMSE | Precision | | Recall | |
|---|---|---|---|---|---|---|
| | | | @10 | @20 | @10 | @20 |
| RMF | .630 | .810 | .682 | .559 | .629 | .803 |
| DMF | .618 | .805 | .671 | .569 | .631 | .810 |
| CDL | .616 | .802 | .672 | .568 | .633 | .810 |
| MDA | .621 | .816 | .680 | .555 | .620 | .802 |
| HLFM | Does not run at this scale | | | | | |
| **40-20** | **.613** | **.802** | **.689** | **.569** | **.632** | **.813** |
| **40-20-10** | **.600** | **.794** | **.696** | **.579** | **.640** | **.820** |

# 5. Conclusion

This is the initial work that introduces the deep latent factor model (deepLFM). We have compared with other deep and shallow techniques and shown that the proposed method outperforms all; at least on the benchmark databases compared on.

At a later stage we would like to incorporate neighbourhood information from the users and items in a graph based deep latent factor framework. This will borrow ideas from graph signal processing [35, 36]. The basic idea would be to regularize the proposed model with trace of graph Laplacians in a fashion similar to graph regularized matrix factorization [36]. These graph Laplacians will be defined from the similarities of the users and the items.

In recent years, tensor decomposition / factorization has been gaining importance in signal processing and machine learning [37, 38]. The way we do multi-level factorizations in this work, it would be interesting to introduce multi-level tensor factorization in the future.

Collaborative filtering / recommender systems have benefitted from the use of additional user-item metadata that has been incorporated into the matrix completion framework via graph regularization [39, 40]. In the future, we would like to improve our method in a similar fashion using graph regularization.

## Acknowledgement

This work is supported by the Infosys Center for Artificial Intelligence @ IIIT Delhi and by the Indo-French CEFIPRA grant DST-CNRS-2016-02.